\documentclass{article}
\usepackage[final]{nips_2016}
\usepackage[utf8]{inputenc} 
\usepackage[T1]{fontenc}    
\usepackage{hyperref}       
\usepackage{url}            
\usepackage{booktabs}       
\usepackage{amsfonts}       
\usepackage{nicefrac}       
\usepackage{microtype}      
\usepackage{graphicx}
\usepackage{graphics}
\usepackage{multicol}
\usepackage{caption}
\usepackage{subcaption}
\usepackage{multicol}
\usepackage{float}
\usepackage{amsmath}

\usepackage{color}
\usepackage{ulem}

\title{Towards deep learning with spiking neurons in energy based models with
contrastive Hebbian plasticity}

\author{
  Thomas Mesnard\\
  \'{E}cole Normale Sup\'{e}rieure\\
  Paris, France \\
  \texttt{thomas.mesnard@ens.fr} \\
   \And
   Wulfram Gerstner \\
   \'{E}cole Polytechnique F\'{e}d\'{e}ral de Lausanne\\
   1015 Lausanne, Switzerland \\
   \texttt{wulfram.gerstner@epfl.ch} \\
   \AND
   Johanni Brea \\
   \'{E}cole Polytechnique F\'{e}d\'{e}ral de Lausanne\\
   1015 Lausanne, Switzerland \\
   \texttt{johanni.brea@epfl.ch} \\
}

\begin{document}
\DeclareGraphicsExtensions{.pdf,.png,.jpg,.svg}

\maketitle

\begin{abstract}
	In machine learning, error back-propagation in multi-layer neural networks
	(deep learning) has been impressively successful in supervised and
	reinforcement learning tasks. As a model for learning in the brain, however,
	deep learning has long been regarded as implausible, since it relies in its
	basic form on a non-local plasticity rule. To overcome this problem,
	energy-based models with local contrastive Hebbian learning were proposed
	and tested on a classification task with networks of rate neurons. We
	extended this work by implementing and testing such a  model with networks
	of leaky integrate-and-fire neurons. Preliminary results indicate that it is
	possible to learn a non-linear regression task with hidden layers, spiking
	neurons and a local synaptic plasticity rule. 
\end{abstract}

\section{Introduction}

Error back-propagation is an efficient and vastly used algorithm to train deep
neural networks \cite{LeCun2015}. Although the architecture of deep learning was
inspired by biological neural networks, adaptation of this powerful algorithm
for training deep \emph{spiking} neural networks started to appear only recently
\cite{Bengio2016,Guergiuev2016}.  

There are several reasons why we are still waiting for a full ``integration of
deep learning and neuroscience'' \cite{Marblestone2016}. In standard deep
learning with feed-forward networks, rate-based neurons with real-valued,
non-negative output are layer-wise sequentially activated to compute the
prediction of the network given a certain input. After comparing this prediction
with a target value, errors are back-propagated in the reversed order through
the layers, using the transposed of the feed-forward weight matrices, but not
the non-linearity of the neurons used in the forward pass to compute the
prediction. The update of weights and biases depends both on the neural
activations of the forward pass and the back-propagated errors. There is no
notion of physical time in standard deep learning. In contrast, biological
neural networks consist of spiking neurons with binary outputs that work in
continuous time. It is unclear how such a network composed of real spiking
neurons could switch between non-linear forward passes and linear backward
passes to implement error back-propagation. 

Equilibrium propagation was recently proposed by Scellier and Bengio
\cite{Scellier2016} as one way to introduce physical time in deep learning and
remove the necessity of different dynamics in forward and backward pass. Their
work combines related ideas like recurrent back-propagation \cite{Pineda1987},
infinitesimal perturbation of the outputs \cite{Oreilly1996,Hertz1997} or
contrastive Hebbian learning \cite{Movellan1991,Xie2003} and moves these
concepts closer to biology.  In a recurrent network of rate-based neurons, whose
dynamics is defined by following the negative gradient of an energy function,
Scellier and Bengio propose to relax the network to a minimum energy state in
the forward phase while fixing the rate of the input neurons at a given
value. The rate of the output neurons at the fixed point of the forward phase
corresponds to the prediction in standard deep learning. Moving the rate of the
output neurons in direction of the target value in the backward phase while
keeping the input rates fixed, perturbs also the rate of the hidden neurons, if
backward connections exist. The key result of Scellier and Bengio is that
supervised learning with this network is possible with a simple contrastive
Hebbian plasticity mechanism that subtracts the correlation of firing rates at
the fixed point of the forward phase from the correlation of the firing rates
after perturbation of the output.

Our contribution is to implement equilibrium propagation in a network of
integrate-and-fire neurons and test it on a regression problem. Our model
differs slightly from the rate-based model in that neither the input rates get
explicitly fixed nor the output rates explicitly moved in direction of the
target. Instead, the input neurons receive a constant input current during both
phases and the output neurons are treated as two-compartment neurons that receive
an extra somatic input current in the backward phase \cite{Urbanczik2014}. The
contrastive Hebbian plasticity mechanism is implemented with an estimate of the
firing rate by low-pass filtering the spike history with a large time constant.

\section{Derivation of the learning rule for rate-based neurons}
For completeness, we reproduce here the learning rule derivation described by
Scellier and Bengio \cite{Scellier2016}.

Let us define the dynamics of a neural network by an energy function
\begin{align}
  2E(s; \hat s, \theta, \beta_x, \beta_y) = 
    &\sum_is_i^2 - 
    \sum_{i\neq j}w_{ij}\rho(s_i)\rho(s_j) -
    2\sum_ib_i\rho(s_i)\,  + \nonumber\\
    &+\, \beta_x \sum_{i\in\mathcal X}(\hat s_i - s_i)^2 + 
    \beta_y \sum_{i\in\mathcal Y}(\hat s_i - s_i)^2\, ,\label{eq:energy}
\end{align}
where $s_i$ is the state of neuron $i$, $\rho(s_i)$ its firing rate for some
non-linear function $\rho$, parameters $\theta = (w, b)$, with
symmetric connection strengths $w_{ij}=w_{ji}$ and biases $b_i$,
$\mathcal Y$ and $\mathcal X$ are disjoint subsets of neurons that may receive
external input $\hat s_i$, if $\beta_x > 0$ or $\beta_y > 0$. Note that the
network does not need to have all-to-all connections. By setting some of the
weights $w_{ij}$ to zero, a multi-layer architecture can be achieved.

The neural dynamics is given by 
\begin{align}
  \tau \dot  s_i &= -\frac{d}{ds_i}E( s; \hat s, \theta,\beta_x, \beta_y)
  \nonumber\\
  &= -s_i +
  \rho'(s_i)\left(\sum_{j}w_{ij}\rho(s_j) + b_i\right)+ 
  \mathbb I_{i\in\mathcal X}\beta_x(\hat s_i - s_i) +
  \mathbb I_{i\in\mathcal Y}\beta_y(\hat s_i - s_i) 
  \label{eq:dynamics}\, ,
\end{align}
with time constant $\tau$ and we used the indicator function $\mathbb
I_{i\in\mathcal Y} = 1$ if $i$ is in set $\mathcal Y$ and 0 otherwise. Note that
for the rectified-linear function, i.e. $\rho(s) = s$ if $s>0$ and $\rho(s) =0$
otherwise, the derivative $\rho'(s) = 1$ for $s>0$. Since for negative $s$ the
derivative $\rho'(s) = 0$, $s$ remains non-negative all the time.

The lowest energy state $s^*$ given by
\begin{equation}
  \frac{dE}{ds_i}(s^*; \hat s, \theta, \infty, 0) = 0, \forall i
  \label{eq:lowestenergystate}
\end{equation}
defines a map $x\mapsto y$, with $x = (s^*_{i_1}, \ldots, s^*_{i_{N_x}})_{i_k
\in\mathcal X}$ and $y = (s^*_{i_1}, \ldots, s^*_{i_{N_y}})_{i_k \in\mathcal
Y}$. We would like to have a rule for changing the parameters $w$ and $b$ to
implement an arbitrary map from $x$ to $y$.

Let us define a cost function for a single pair of points $\hat s$ and $s$
\begin{equation}                                                         
  2C(\hat s, s) = \sum_{i\in\mathcal Y}(\hat s_i - s_i)^2
\end{equation}
and a total cost function $C(\hat s^1, s^1, \ldots, \hat s^N, s^N) =
\frac1N\sum_{\mu = 1}^NC(\hat s^\mu, s^\mu)$. 

To find a learning rule, we look at the constraint optimization problem
\begin{equation}
  \min_{\theta} C(\hat s^1, s^1, \ldots, \hat s^N, s^N) \mbox{ subject to }
  \frac{dE}{ds_i}(s^\mu; \hat s^\mu, \theta, \infty, 0) = 0, \forall \mu, i
\end{equation}
and define the Lagrangian for a single data point $\hat s$ by
\begin{equation}
  \mathcal L(s, \lambda, \theta; \hat s) = C(\hat s, s) + \sum_i\lambda_i
  \frac{dE}{ds_i}(s; \hat s,\theta,  \infty, 0)
\end{equation}

We minimize this Lagrangian by setting the derivatives with respect to $s$ and
$\lambda$ to zero 
\begin{align}
  \frac{d\mathcal L}{d\lambda_i}(s^*, \lambda^*, \theta; \hat s) = 0 , \forall
  i\label{eq:constraint1}\\
  \frac{d\mathcal L}{ds_i}(s^*, \lambda^*, \theta; \hat s) = 0, \forall
  i\label{eq:constraint2}
\end{align}
and performing stochastic descent on the total cost by changing
the parameters according to
\begin{align}
  \Delta \theta_i &= - \eta \frac{\partial \mathcal L
	}{\partial\theta_i}(s^*,\lambda^*,\theta; \hat s)\\
  &= -\eta\sum_j\lambda^*_j\frac{\partial dE}{\partial\theta_ids_j}(s^*; \hat s, \theta, \infty,
  0) \label{eq:lruleabstract}
\end{align}
where $\eta$ is a learning rate. 

Solving \autoref{eq:constraint1}, we find that the state $s^*$ is simply given
by the lowest energy state (c.f.  \autoref{eq:lowestenergystate}), i.e. it can
be obtained by running the dynamics with input $x$ ($\beta_x = \infty$) and no
target y ($\beta_y = 0$).

\autoref{eq:constraint2} is a bit harder to solve for $\lambda^*$. Expanding the
definition of the Lagrangian it reads
\begin{equation}
  \frac{dC}{ds_i}(\hat s, s^*) + \sum_j\lambda^*_j\frac{d^2E}{ds_ids_j}(
  s^*; \hat s,\theta,\infty, 0) = 0\, .\label{eq:constraint2explicit}
\end{equation}
The nice idea of Scellier and Bengio \cite{Scellier2016} is, to look at the fixed point $s^\beta$ of the
dynamics with weak target input $\beta_y = \beta > 0$ given by
\begin{equation}
  \frac{dE}{ds_i}(s^\beta; \hat s,\theta,  \infty, \beta) = 0\, ,\forall
  \beta\, .
\end{equation}
Since the left-hand side of this equation is a constant function in $\beta$,
its derivative with respect to $\beta_y$ is also zero and evaluated at $\beta_y
= 0$ we get
\begin{align}
  0=\frac{d^2E}{d\beta_yds_i} &= 
  \frac{\partial dE}{\partial\beta_yds_i} +
  \sum_j\frac{ds^\beta_j}{d\beta_y}\frac{d^2E}{ds_ids_j}\\
  &=\frac{dC}{ds_i}(\hat s, s^*) +
  \sum_j\frac{ds^\beta_j}{d\beta_y}\frac{d^2E}{ds_ids_j}(s^*; \hat s,\theta, 
  \infty, 0)\, ,
\end{align}
which has the same form as \autoref{eq:constraint2explicit} if we identify
$\lambda^*_j = \frac{ds^\beta_j}{d\beta_y}$.

Using this result in \autoref{eq:lruleabstract}, we can write
    \begin{align}
      \sum_k\lambda_k^*\frac{\partial dE}{\partial w_{ij}ds_k} & = 
      \sum_k \frac{ds^\beta_k}{d\beta_y}\frac{d\partial E}{ds_k\partial
      w_{ij}} =\frac{d\partial E}{d\beta_y\partial w_{ij}} =
      -\frac{d}{d\beta_y}\big(\rho(s_i)\rho(s_j)\big)
    \end{align}
    and approximate this by $\frac{d}{d\beta_y}\big(\rho(s_i)\rho(s_j)\big) 
	\approx \frac1\beta\big(\rho(s_i^\beta) \rho(s_j^\beta) -
	\rho(s_i^*)\rho(s_j^*)\big)$ which
      leads to the contrastive Hebbian update rule
      \begin{equation}
        \Delta w_{ij} = \eta \big(\rho(s_i^\beta)\rho(s_j^\beta) -
        \rho(s_i^*) \rho(s_j^*)\big)\, ,\label{eq:contrastivehebb}
      \end{equation}
where the factor $1/\beta$ is absorbed into the learning rate $\eta$.

This learning rule can be implemented by iterating through:
\begin{enumerate}
	\item Select a data sample $\hat s$ and relax the system to the lowest energy
		state with $\beta_y = 0$ to obtain $y^*$ (Forward phase).
	\item Subtract $\eta\rho(s_i^*)\rho(s_j^*)$ from the weights.
	\item Set $\beta_y=\beta>0$ and let the system evolve for some
		time (Backward phase).\footnote{Scellier and Bengio \cite{Scellier2016} observe that
			relaxation to the fixed-point is not necessary in this second
		phase.}
	\item Add $\eta\rho(s_i^\beta)\rho(s_j^\beta)$ to the weights.
\end{enumerate}

\section{Implementation with leaky integrate-and-fire neurons}

\begin{figure}[t]
	\begin{tabular}{p{3mm}p{9cm}p{3mm}p{3cm}}
	\textbf{A}&\vspace{0pt} 
  \includegraphics[height=.22\textheight]{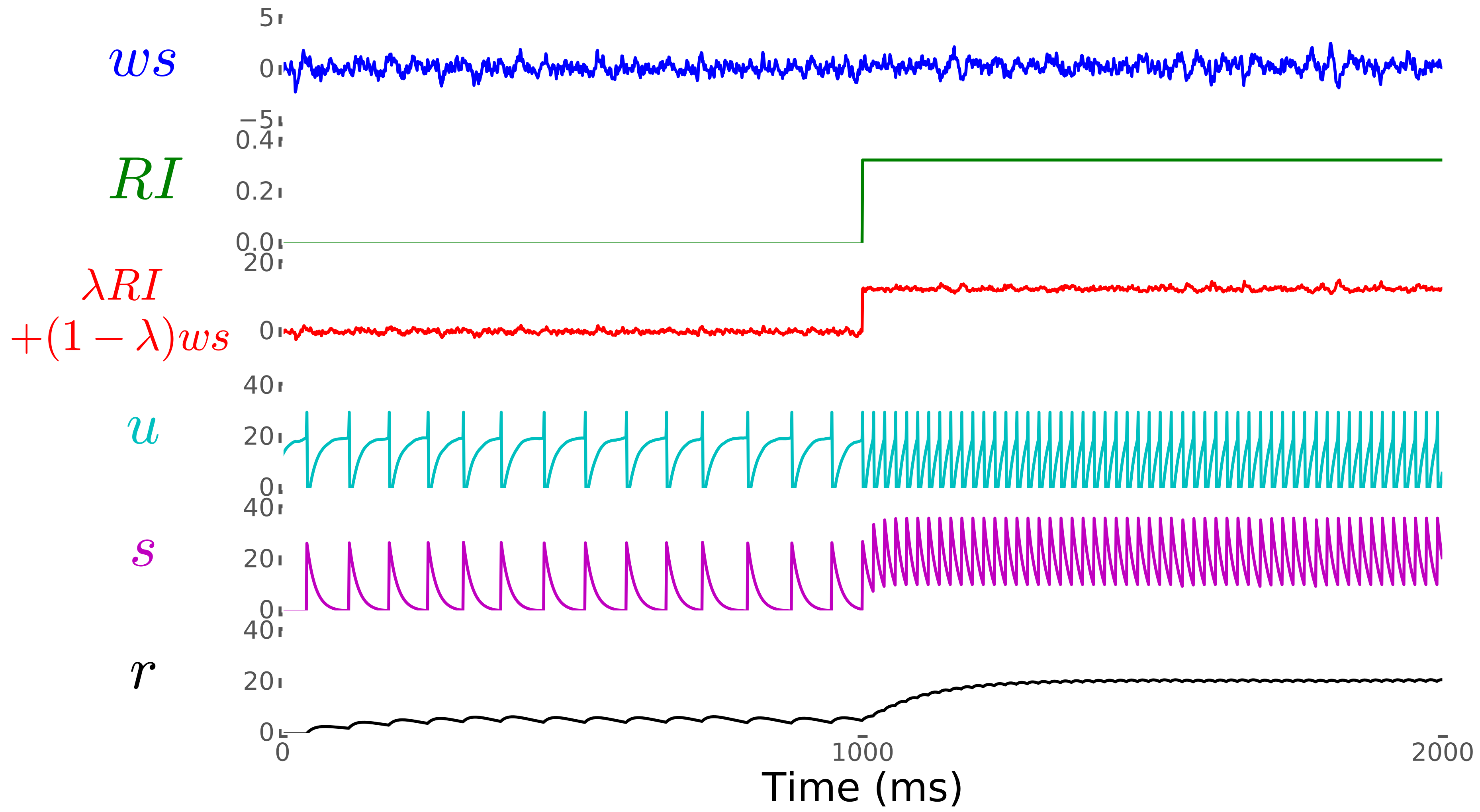}
  &\textbf{B}&\vspace{0pt} 
	\includegraphics[height=.22\textheight]{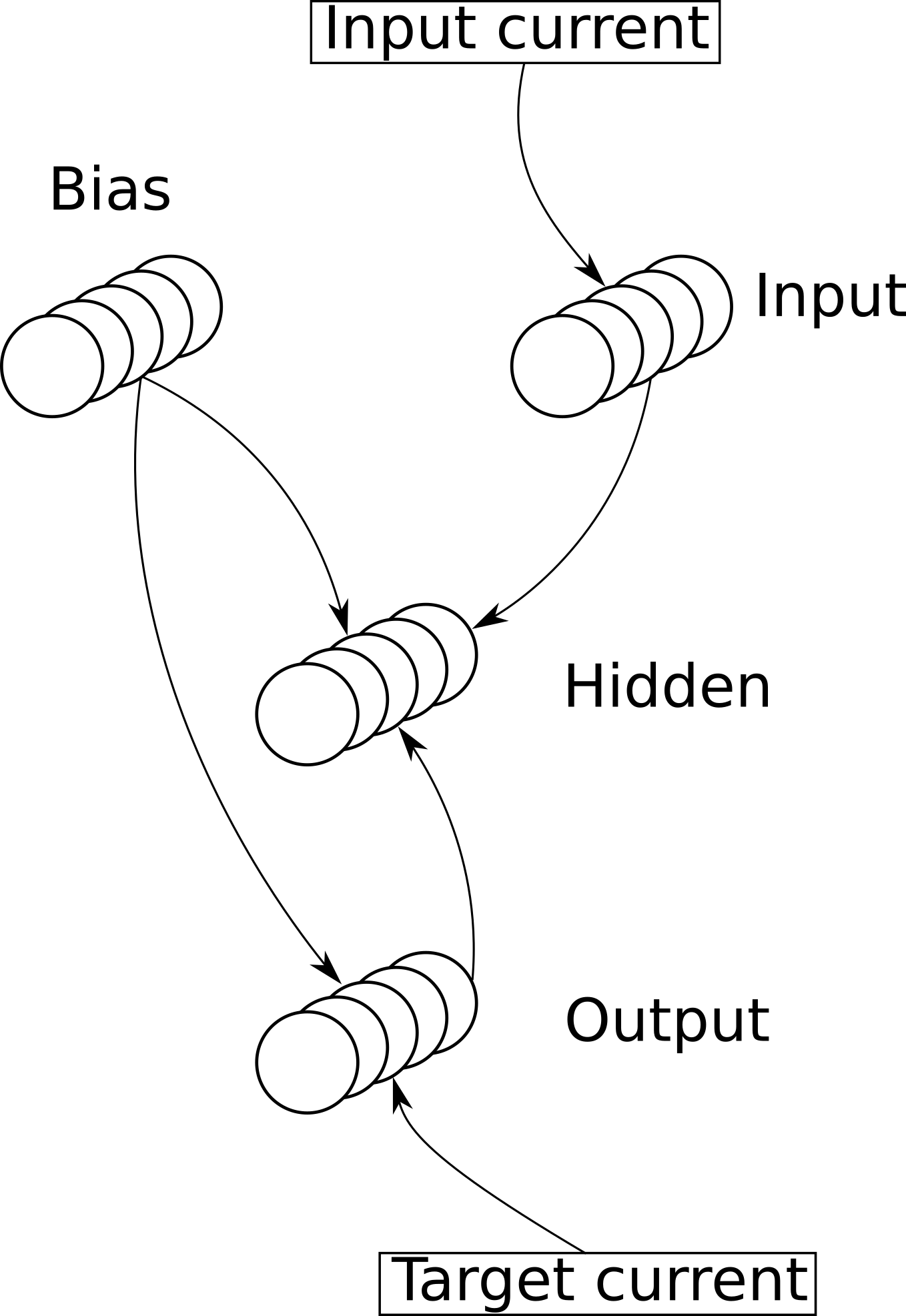}
\end{tabular}
	\caption{\textbf{Neuron model and network architecture. A} Traces of the
		relevant quantities for a target neuron. The synaptic weights are
		updated according to 
		$\Delta w_{ij}=\eta (r_i^+r_j^+-
		r_i^-r_j^-)$, where $r_i^+=r_i(1200\,\mathrm{ms})$ and
		$r_i^-=r_i(600\,\mathrm{ms})$,
		which can be implemented online by decreasing the weights
		appropriately at 600 ms and increase them at 1200 ms. \textbf{B} Network
		architecture. Arrows indicate all-to-all connectivity. }
		\label{fig:model}
\end{figure}

To replace rate-based neurons with leaky integrate-and-fire neurons we introduce the
somatic membrane potential $u_i$ of neuron $i$ that evolves below a threshold
$\theta$ as
\begin{align}
	\tau \dot u_i(t) &= -u_i(t) + u_0 + \big(1 - \lambda_i(t)\big)\sum_j w_{ij}
	s_j(t) + \lambda_i(t)
	RI_i(t)\label{eq:leakyintegrateandfire}
\end{align}
with time constant $\tau$, resting potential $u_0$, post-synaptic response $s_j$
given by the dynamics $\tau_s\dot s_j = -s_j + u_{psp}x_j$, with presynaptic
spike trains $x_j(t) = \sum_{t_j^{(f)}}\delta(t - t_j^{(f)})$, where $t_j^{(f)}$
are the spike times of neuron $j$, membrane resistance $R$ and additional
current input $I_i(t)$, used in the backward phase to nudge the firing rate of
the neuron in direction of the target firing rate.  The nudging factor 
\begin{equation}
	\lambda_i(t) = \frac{RI_i(t)}{RI_i(t) + \sum_j w_{ij}
	s_j(t)}\, ,
\end{equation}
is important at the end of learning, when the predictions by the network match
almost the target inputs and additive instead of convex combination of network
input and target input would lead to run-away dynamics. The nudging factor can
be motivated with divisive normalization \cite{Carandini2012} or an argument
involving conductance-based synapses \cite{Urbanczik2014}. The neuron spikes
when its membrane potential reaches threshold $\theta$. The membrane potential
is then set to a reset value $u_r$ and kept at this value for a refractory
period $\Delta$. Afterwards the dynamics in Eq.~\ref{eq:leakyintegrateandfire}
determines again the membrane potential.

To implement the contrastive Hebbian learning rule in \autoref{eq:contrastivehebb},
the pre- and postsynaptic firing rates are estimated in each synapse with 
hypothesized processes $r_i$ that low-pass filter the
spike trains with a large time constant $\tau_r$, i.e.
\begin{equation}
	\tau_r\dot r_i(t) = -r_i(t) + s_i(t)\, .
\end{equation}
The negative part of the weight update in \autoref{eq:contrastivehebb} is
applied just before the network receives target input to the output neurons. The
positive part of the weight update is applied when the network evolved for some
time under the influence of the target input.

\autoref{fig:model}A shows a trace of the relevant quantities for one target
neuron. 

The firing rate $f$ of a leaky integrate-and-fire neuron with dynamics $\tau \dot u
= -u + v$, constant drive $v$, reset potential $u_r$, threshold $\theta$  and refractory period $\Delta$ is given by
\begin{equation}
	f(v) = \frac{1}{\tau \log\left(\frac{v - u_r}{v - \theta}\right) +
	\Delta}\quad\mbox{ if }\quad v>\theta, \quad f(v) = 0\quad\mbox{ otherwise}\, .
  \label{eq:liffi}
\end{equation}
The rate model ``closest'' to the spiking model uses this non-linearity, i.e.
$\rho(v_i) = f(v_i)$ with $v_i = u_0 + \big(1 - \lambda_i(t)\big)\sum_j w_{ij}
	s_j(t) + \lambda_i(t)
	RI_i(t)$. Even though the derivative of this non-linearity is not
piecewise constant as the one of the rectified-linear function, we did not see a
significant difference in simulations when using $\rho'(v) = 1$ for $v>0$ and
$\rho'(v) = 0$ otherwise together with the leaky integrate-and-fire
non-linearity $f$.

For the experiments described in the next section we used a network architecture
with one hidden layer (see \autoref{fig:model}). We used only one hidden layer
because the task is learnable with one, but nothing prevents the use of multiple
hidden layers. Bias terms are implemented with weights from bias neurons
that are active at a constant firing rate. Instead of clamping the input neurons
to the input values with $\beta_x\rightarrow \infty$, we did not allow any
feedback from the other layers to the input layer and provided the input as
input currents $I_i$ to the neurons in the input layer. 

\subsection{Results}
\label{subsection:Results}

\begin{figure}[t]
	\centering
	\includegraphics[width=\textwidth]{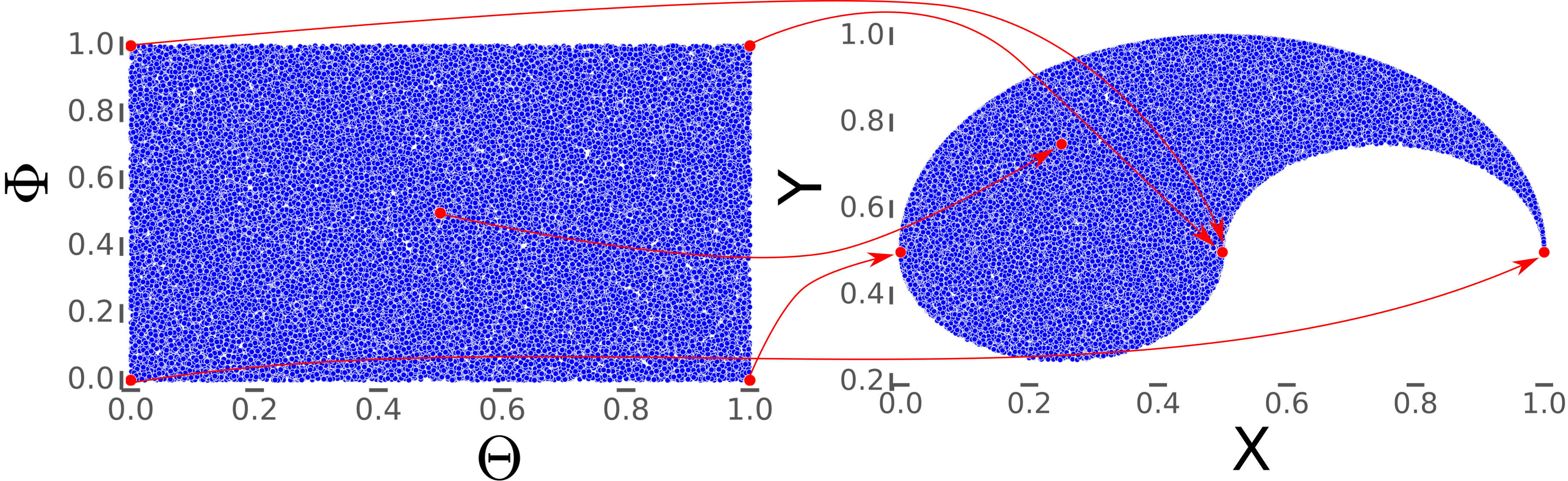}
	\caption{\textbf{A non-linear regression task.} Input values $\theta\in[0,1]$
		and $\varphi\in[0,1]$ are mapped to target values $x =
		\big(\cos(\pi\varphi)+\cos(\pi(\varphi +\theta))+2\big)/4$ and $y =
		\big(\sin(\pi\varphi)+\sin(\pi(\varphi +\theta))+2\big)/4$.}\label{fig:task}
\end{figure}

In contrast to Scellier and Bengio \cite{Scellier2016}, who tested the model on
MNIST classification, we choose a non-linear regression task (see
\autoref{fig:task}). Since energy-based models, like the Hopfield model
\cite{Hopfield1982}, are well known with parameter settings that create multiple
local minima, we were curious to see, if learning is also possible when local
minima are undesirable. 

\subsubsection{Rate-based neurons}
\label{result_firing}
  
\begin{figure}[t]
  \centering
  \begin{minipage}[c]{.49\linewidth}
          \includegraphics[width=\textwidth]{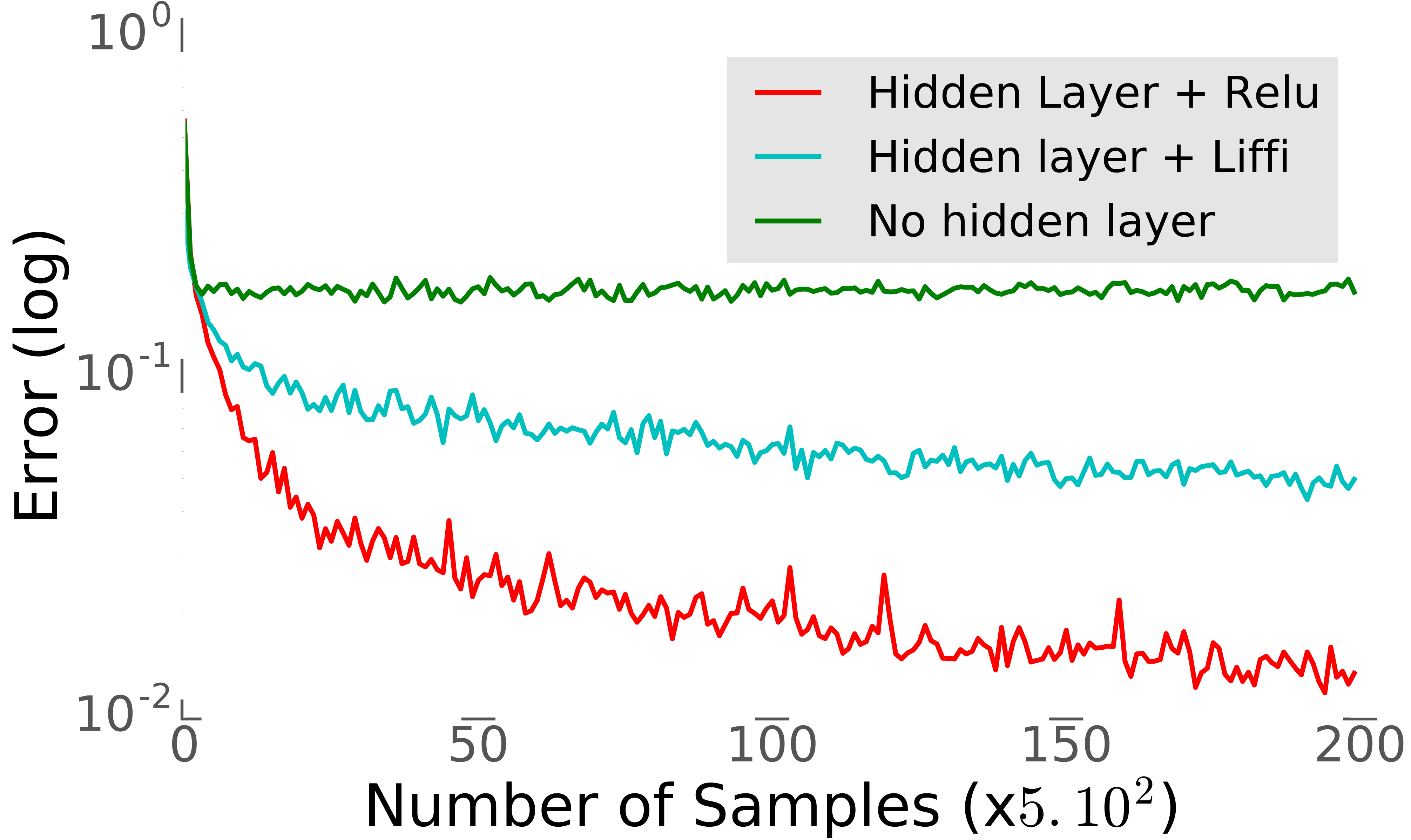}
  \end{minipage} \hfill
  \begin{minipage}[c]{.49\linewidth}
        \includegraphics[width=\textwidth]{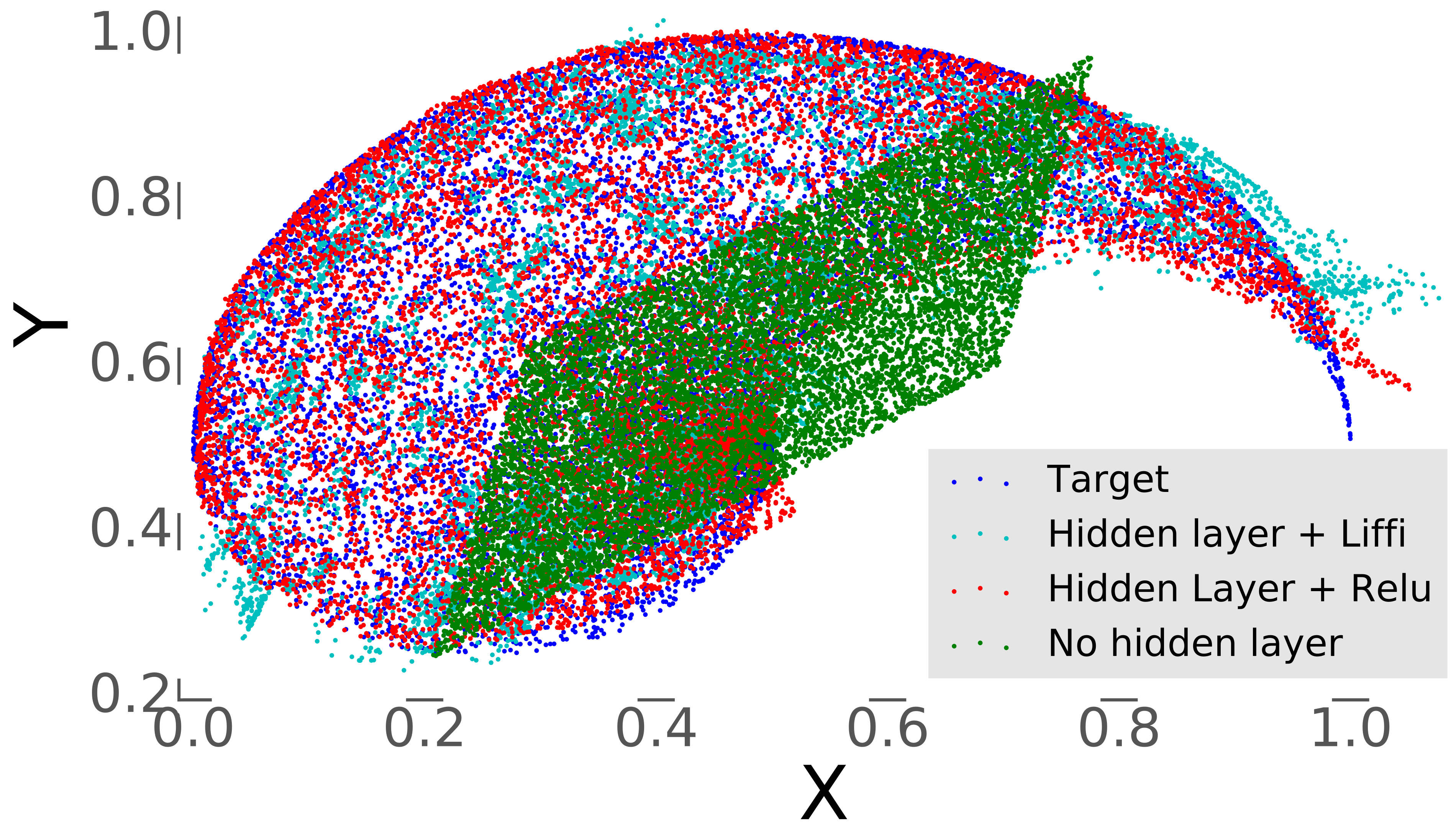}
   \end{minipage}
   \caption{\textbf{Learning the regression task with networks of rate-based
   neurons.} The error is the estimated average Euclidian distance between
   prediction and target. The differential equations were integrated with Euler
   method, time steps 1 ms, $\tau = 15$ ms, $\tau_s=10$ ms, $\tau_r=300$ ms,
   $u_0 = 20$, $u_r=0$, $\theta = 20$, $u_\mathrm{psp}=400$, $R=40$, $\Delta =
   5$ ms, $I_i\in[0,1]$, duration of forward and backward phase 600 ms, constant learning
   rates $\eta_i = 0.1/\sqrt{\mathrm{indegree}_i}$, where indegree$_i$ is the
   number of presynaptic neurons of neuron $i$. Weights are randomly initialized
   to small values without symmetry constraints. Note that the learning rule
   \autoref{eq:contrastivehebb} tends to symmetrize the weight matrix. We tested
   two non-linearities: the rectified linear function (green and red) and the
   Liffi function defined in \autoref{eq:liffi} (cyan). Preliminary results
   (not shown) indicate that learning with Liffi is slower, even when
   learning rates are optimized for each non-linearity. }\label{fig:ratebased}
\end{figure}  

In \autoref{fig:ratebased}, we see that the task can be learned with a single
layer of 400 rate-based neurons. Learning is a bit slower when the firing rate
$f$ of the leaky integrate-and-fire neuron (c.f. \autoref{eq:liffi}) is used
instead of the rectified linear function. As expected, the task cannot be
learned without this hidden layer.

In these simulations, weights were updated all in once after the backward phase
by following \autoref{eq:contrastivehebb}. With small learning rates, the same
results are expected, if the weights are updated online, i.e. after the
forward phase by subtracting $\eta\rho(s_i^*)\rho(s_j^*)$ and after the backward
phase by adding $\eta\rho(s_i^\beta)\rho(s_j^\beta)$.

\subsubsection{Spiking neurons}
\label{subsubsection_Spiking}
In the spiking implementation each real-valued input and output dimensions was
represented by 20 neurons that received identical current inputs and the hidden layer is composed of 300 neurons.  

\begin{figure}[ht]
  \centering
  \begin{minipage}[c]{.49\linewidth}
          \includegraphics[width=\textwidth]{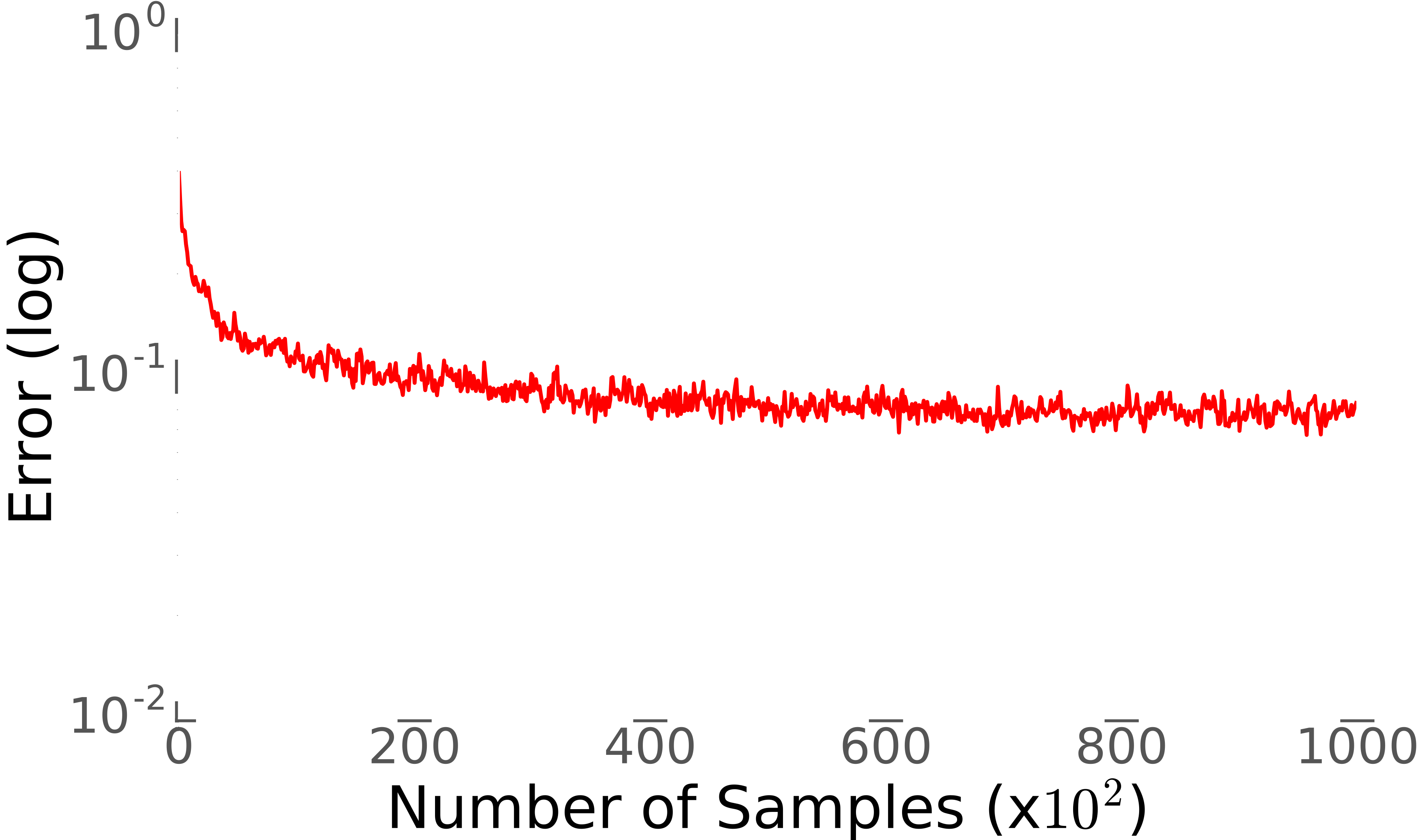}
  \end{minipage} \hfill
  \begin{minipage}[c]{.49\linewidth}
        \includegraphics[width=\textwidth]{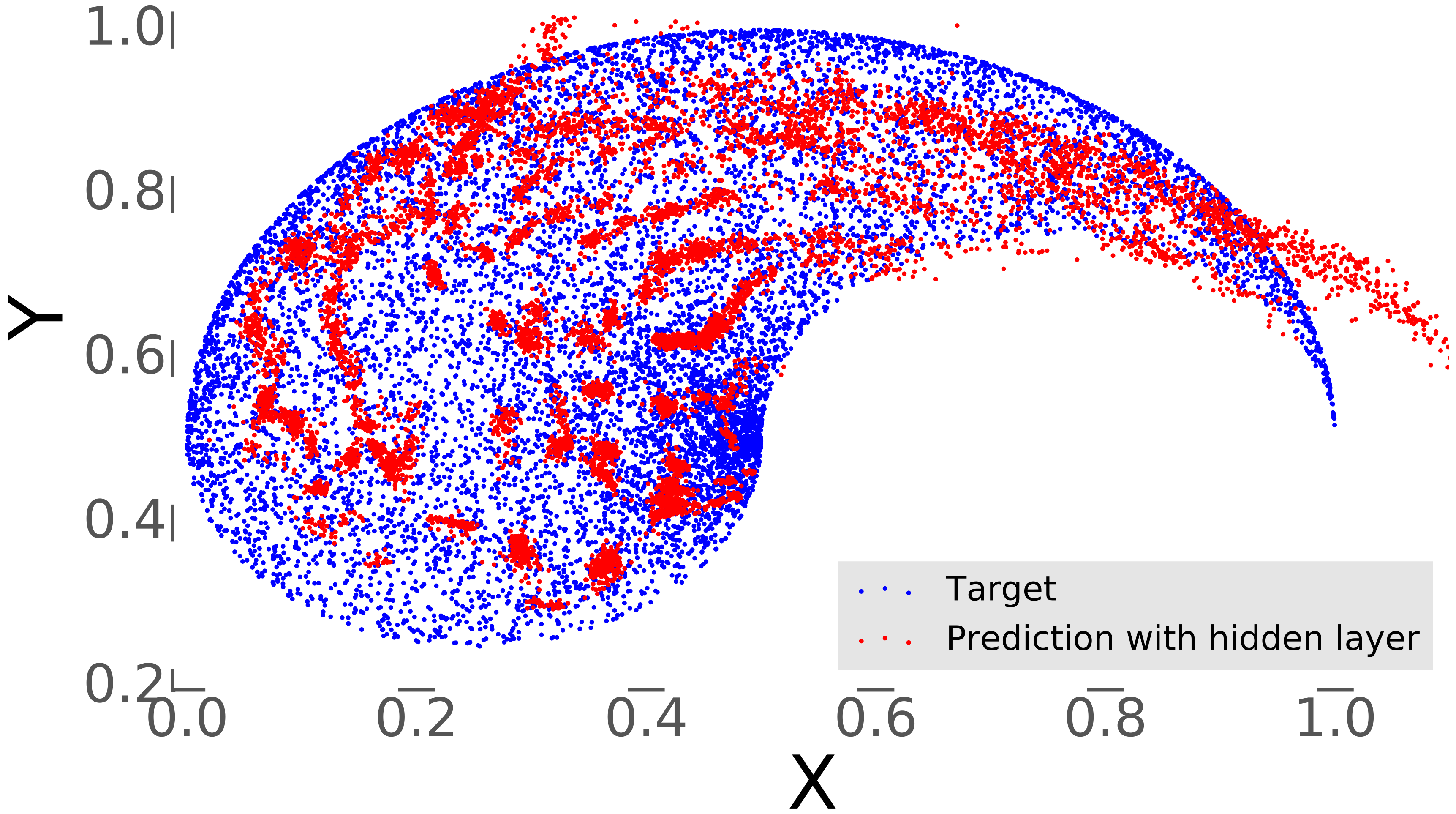}
   \end{minipage}
   \caption{\textbf{Learning the regression task with a network of leaky
   integrate-and-fire neurons.} The output values $x$ and $y$ are given by the
   average firing rate of 20 neurons. The differential equations were integrated with Euler method, time steps 1 ms, $\tau = 15$ ms, $\tau_s=15$ ms, $\tau_r=100$ ms, $u_0 = 20$, $u_r=0$, $\theta = 20$, $u_\mathrm{psp}=400$, $R=40$, $\Delta =
   5$ ms, $I_i\in[0,1]$, duration of forward and backward phase 1000 ms, constant learning
   rates $\eta_i = 5.10^{-5}/\sqrt{\mathrm{indegree}_i}$, where indegree$_i$ is the
   number of presynaptic neurons of neuron $i$.}\label{fig:spiking}
\end{figure}

\autoref{fig:spiking} shows a typical result with spiking neurons. Even though
the target is not perfectly learned, the predictions are much more accurate than
without a hidden layer (c.f. \autoref{fig:ratebased}). In the simulations with
rate-based neurons we noticed the importance of converging close to the fixed
point in the forward phase. The poorer results with spiking neurons could be
partially due to not having a good enough estimate of the fixed point in the forward phase.

\section{Discussion}
We have implemented Equilibrium Propagation \cite{Scellier2016} with a
multilayer network of leaky integrate-and-fire neurons and demonstrated that the
network can learn a non-linear regression task. The results with the spiking
networks are not as convincing as with the rate-based networks, which could be
due to non-optimal estimates of the fixed point in the forward phase.

The locality of the contrastive Hebbian plasticity rule and the error
back-propagation by natural recurrent dynamics make this approach appealing from
the perspective of biological plausibility. The plasticity mechanism requires
quite precisely timed induction of anti-Hebbian and Hebbian plasticity.
Potential processes to implement this involve theta waves \cite{orr2001} or
neuromodulators \cite{pawlak2010timing,fremaux2015neuromodulated}, known for
modulating and even sign-reversing synaptic plasticity.

Equilibrium propagation is not the only attempt to implement error
back-propagation in a more biologically plausible way. A recent and compelling
approach \cite{Guergiuev2016} postulates an implementation of error
back-propagation using the computational richness of multi-compartment neurons.

It is exciting to see the theoretical approaches that try to integrate deep
learning and neuroscience. Ultimately, experiments are needed to support or
falsify these hypotheses.
\section*{Acknowledgments}

\bibliographystyle{unsrt}

\end{document}